\documentclass{ecai2014}
\usepackage{times}
\usepackage{graphicx}
\usepackage{latexsym}
\usepackage{amsmath}
\usepackage{color}

\begin{document}

\title{Recommending Learning Algorithms and\\Their Associated Hyperparameters} 

\author{Michael R. Smith\institute{Brigham Young University, USA, email: msmith@axon.cs.byu.edu} \and Logan Mitchell\institute{Brigham Young University, USA, email: mitchlam711@gmail.com} \and Christophe Giraud-Carrier\institute{Brigham Young University, USA, email: cgc@cs.byu.edu} \and Tony Martinez\institute{Brigham Young University, USA, email: martinez@cs.byu.edu}}

\maketitle
\bibliographystyle{ecai2014}

\begin{abstract}
The success of machine learning on a given task depends on, among other things, which learning algorithm is selected and its associated hyperparameters. Selecting an appropriate learning algorithm and setting its hyperparameters for a given data set can be a challenging task, especially for users who are not experts in machine learning. Previous work has examined using meta-features to predict which learning algorithm and hyperparameters should be used. However, choosing a set of meta-features that are predictive of algorithm performance is difficult. Here, we propose to apply collaborative filtering techniques to learning algorithm and hyperparameter selection, and find that doing so avoids determining which meta-features to use and outperforms traditional meta-learning approaches in many cases.
\end{abstract}

\section{Introduction}

Most previous meta-learning work has focused on selecting a learning algorithm or a set of hyperparameters based on meta-features used to characterize datasets~\cite{Brazdil2009}. As such, it can be viewed as a form of content-based filtering, a technique commonly-used in recommender systems that captures a set of measured characteristics of an item and/or user to recommend items with similar characteristics. On the other hand, collaborative filtering (CF), also used by some recommender systems, predicts the rating or preference that a user would give to an item, based on the past behavior of a set of users, characterized by ratings assigned by users to a set of items~\cite{koren:matrix_factorization}. The underlying assumption of CF is that if users $A$ and $B$ agree on some issues, then user $A$ is more likely to have the same opinion on a new issue $X$ as user $B$ than another randomly chosen user. A key advantage of CF is that it does not rely on directly measurable characteristics of the items. Thus, it is capable of modeling complex items without actually understanding the items themselves. 

Here, we propose \textit{meta-CF} (MCF) a novel approach to meta-learning that applies CF in the context of algorithm and/or hyperparameter selection. MCF differs from most previous meta-learning techniques in that it does not rely on meta-features. Instead, MCF considers the similarity of the performance of the learning algorithms with their associated hyperparameter settings from previous experiments. In this sense, the approach is more similar to landmarking~\cite{Pfahringer2000} and active testing~\cite{Leite2012} since both also use the performance results from previous experiments to determine similarity among data sets.

While algorithm selection and hyperparameter optimization have been mostly studied in isolation (e.g., see~\cite{Pfahringer2000,Brazdil2003,Ali2006a,Ali2006b,Bergstra2012,Snoek2012}), recent work has begun to consider them in tandem. For example, Auto-WEKA simultaneously chooses a learning algorithm and sets its hyperparameters using Bayesian optimization over a tree-structured representation of the combined space of learning algorithms and their hyperparameters~\cite{Thornton2013}. All of these approaches face the difficult challenge of determining a set of meta-features that capture relevant and predictive characteristics of datasets. By contrast, MCF does consider both algorithm selection and hyperparameter setting at once, but alleviates the problem of meta-feature selection by leveraging information from previous experiments through collaborative filtering.

Our results suggest that using MCF for learning algorithm/hyperparameter setting recommendation is a promising direction.
Using MCF for algorithm recommendation has some differences from the traditional CF used for human ratings. For example, CF for humans may have to deal with concept drift, where a user's taste may change over time; working with learning algorithms and hyperparameter settings is deterministic.

\section{Empirical Evaluation}

For MCF, we examine several CF techniques implemented in the Waffles toolkit~\cite{Gashler2011}: baseline (predict the mean of the previously seen results), 
Fuzzy K-Means (FKM)~\cite{li_fuzzy_k_means}, Matrix Factorization (MF)~\cite{koren:matrix_factorization}, Nonlinear PCA (NLPCA)~\cite{nlpca}, and Unsupervised Backpropagation (UBP)~\cite{Gashler2014}.

To establish a baseline, we first calculate the accuracy on a set of 125 data sets and 9 diverse learning algorithms (see~\cite{Smith2012_IH} for a discussion on diversity) with default parameters as set in Weka~\cite{weka2009}. The set of learning algorithms is composed of backpropagation (BP), C4.5, $k$NN, locally weight learning (LWL), na\"{i}ve Bayes (NB), nearest neighbor with generalization (NNge), random forest (RF), ridor (Rid), and RIPPER (RIP). We select the accuracy from the learning algorithm that produces the highest classification accuracy. This represents algorithm selection with perfect recall. We also estimate the hyperparameter optimized accuracies for each learning algorithm using random hyperparameter optimization~\cite{Bergstra2012}. The results are shown in Table \ref{table:oracle}, where the accuracy from each learning algorithm is the average hyperparameter optimized accuracy for each data set, ``Default'' refers to the best accuracy from the learning algorithm with its default parameters, ``ALL'' refers to the accuracy from the best learning algorithm and hyperparameter setting, and ``AW'' refers to the results from running Auto-WEKA. For Auto-WEKA, each dataset was allowed to run as long as the longest algorithm took to run on the dataset when doing the random hyperparameter optimization. As Auto-WEKA is a random algorithm, we ran 4 runs each time with a different seed and chose the seed with highest accuracy. This can be seen as equivalent to allowing a user to run on average 16 learning algorithm and hyperparameter combinations on a data set.

\begin{table}
\begin{center}
{\caption{Average accuracy for the best hyperparameter setting for each learning algorithm, algorithm selection (Default), both algorithm selection and hyperparameter optimization (ALL), and Auto-WEKA (AW).}
\label{table:oracle}}
\begin{tabular}{cccccc}
BP & C4.5 & $k$NN & LWL & NB & NNge \\
\hline
\\[-6pt]
79.89 & 79.22 & 78.05 & 77.48 & 76.04 & 76.80 \\
\hline
\\[-6pt]
\\
RF & Rid & RIP & Default & ALL & AW \\
\hline
\\[-6pt]
79.58 & 71.48 & 77.31 & 81.93 & 83.00 & 82.00\\
\hline
\end{tabular}
\end{center}
\end{table}

For MCF, we compiled the results from hyperparameter optimization. We randomly removed 10\% to 90\% of the results by increments of 10\% and then used MCF to fill in the missing values. The top 4 learning algorithm/hyperparameter configurations are returned by the CF technique and the accuracy from the configuration that returns the highest classification accuracy is used. This process was repeated 10 times. A summary of the average results for MCF are provided in Table \ref{table:CF}.
The columns ``Best'', ``Median'', and ``Average'' refer to the accuracies averaged across all of the sparsity levels for the hyperparameter setting for the CF technique that provided the results.
The columns 0.1 to 0.9 refer to the percentage of the results used for CF averaged over the hyperparameter settings.
The row ``Content'' refers to meta-learning where a learning algorithm recommends a learning algorithm based on a set of meta-features.

\begin{table}
\begin{center}
{\caption{Average accuracy from the best of the top 4 recommended learning algorithm and hyperparameter settings from MCF.}\label{table:CF}}
\setlength{\tabcolsep}{3.5pt}
\begin{tabular}{l|ccc||ccccc}
 & Best & Med & Ave & 0.1 & 0.3 & 0.5 & 0.7 & 0.9 \\
\hline
\\[-6pt]
Baseline & 81.11 & 81.11 & 81.11 & \textbf{80.49} & 80.91 & 81.12 & 81.33 & 81.54 \\
FKM & 81.52 & 81.04 & 81.29 & 80.13 & 80.65 & 81.07 & 81.45 & 81.88 \\
MF & \textbf{82.12} & \textbf{82.06} & \textbf{81.95} & \textbf{80.49} & \textbf{81.63} & \textbf{82.12} & \textbf{82.44} & \textbf{82.65} \\
NLPCA & 81.73 & 81.33 & 81.33 & 79.98 & 80.58 & 81.43 & 82.08 & 82.61 \\
UBP & 81.73 & 81.27 & 81.31 & 80.05 & 80.51 & 81.34 & 82.05 & 82.61 \\
\hline
\\[-6pt]
Content & 81.35 & 80.47 & 78.91 & - & - & - & - & -\\
\end{tabular}
\end{center}
\end{table}

Overall, MF achieves the highest accuracy values. The effectiveness of MCF increases as the percentage of the results increases.
MCF significantly increases the classification accuracy compared with both hyperparameter optimization for a given learning algorithm and model selection with their default parameters as well as using the meta-features to predict which learning algorithm and hyperparameters to use.
On average, MCF and Auto-WEKA achieve similar accuracy, which highlights the importance of considering {\em both} algorithm selection and hyperparameter optimization.
However, provided one has access to a database of experiments, such as the ExperimentDB~\cite{Vanschoren2012}, MCF only requires the time to run a number of algorithms (often ran in parallel), and retraining the collaborative filter.
In the current implementation, retraining takes less than 10 seconds.
Thus, MCF presents an efficient method for recommending a learning algorithm and its associated hyperparameters.


While our results show that MCF is a viable technique for recommending learning algorithms {\em and} hyperparameters, some work remains to be done. 
Future work for MCF includes addressing the cold-start problem which occurs when a data set is presented and no learning algorithm has been ran on it.
MCF is adept at exploiting the space that has already been explored, but (like active testing) it does not explore unknown spaces at all. One way to overcome this limitation would be to use a hybrid recommendation system that combines content-based filtering and MCF.

\end{document}